\documentclass[sigconf]{acmart}

\usepackage{graphicx}
\usepackage{amsmath}
\usepackage{bm} 
\usepackage{subcaption}
\usepackage{multirow}
\usepackage{varwidth}
\usepackage{pifont}
\usepackage{makecell}
\usepackage{wrapfig}
\usepackage{diagbox}
\usepackage{tabularx}
\usepackage{xspace}

\AtBeginDocument{%
  \providecommand\BibTeX{{%
    \normalfont B\kern-0.5em{\scshape i\kern-0.25em b}\kern-0.8em\TeX}}}

\copyrightyear{2023}
\acmYear{2023}
\setcopyright{acmlicensed}\acmConference[KDD '23]{Proceedings of the 29th ACM SIGKDD Conference on Knowledge Discovery and Data Mining}{August 6--10, 2023}{Long Beach, CA, USA}
\acmBooktitle{Proceedings of the 29th ACM SIGKDD Conference on Knowledge Discovery and Data Mining (KDD '23), August 6--10, 2023, Long Beach, CA, USA}
\acmPrice{15.00}
\acmDOI{10.1145/3580305.3599898}
\acmISBN{979-8-4007-0103-0/23/08}

\begin{document}

\title{Revisiting Personalized Federated Learning: Robustness Against Backdoor Attacks}

\author{Zeyu Qin\textsuperscript{*}}
\affiliation{\institution{Hong Kong University of Science and Technology
}
   \country{}}
\email{zeyu.qin@connect.ust.hk}

\author{Liuyi Yao}
\affiliation{\institution{Alibaba Group}
   \country{}}
\email{yly287738@alibaba-inc.com}

\author{Daoyuan Chen}
\affiliation{\institution{Alibaba Group}
   \country{}}
\email{daoyuanchen.cdy@alibaba-inc.com}
\author{Yaliang Li}
\affiliation{\institution{Alibaba Group}
   \country{}}
\email{yaliang.li@alibaba-inc.com}

\author{Bolin Ding}
\affiliation{\institution{Alibaba Group}
   \country{}}
\email{bolin.ding@alibaba-inc.com}

\author{Minhao Cheng}
\affiliation{\institution{Hong Kong University of Science and Technology
}
   \country{}}
\email{minhaocheng@ust.hk}

\renewcommand{\shortauthors}{Zeyu Qin et al.}

\begin{abstract}
In this work, besides improving prediction accuracy, we study whether personalization could bring robustness benefits to backdoor attacks. We conduct the first study of backdoor attacks in the pFL framework, testing 4 widely used backdoor attacks against 6 pFL methods on benchmark datasets FEMNIST and CIFAR-10, a total of 600 experiments. The study shows that pFL methods with partial model-sharing can significantly boost robustness against backdoor attacks. In contrast, pFL methods with full model-sharing do not show robustness. To analyze the reasons for varying robustness performances, we provide comprehensive ablation studies on different pFL methods. Based on our findings, we further propose a lightweight defense method, Simple-Tuning, which empirically improves defense performance against backdoor attacks. We believe that our work could provide both guidance for pFL application in terms of its robustness and offer valuable insights to design more robust FL methods in the future. We open-source our code to establish the first benchmark for black-box backdoor attacks in pFL: \textit{\url{https://github.com/alibaba/FederatedScope/tree/backdoor-bench}}.
\end{abstract}



\maketitle

\renewcommand{\thefootnote}{\fnsymbol{footnote}}
\footnotetext[1]{Work was done when the first author Zeyu Qin was an intern at Alibaba Group.} 
\renewcommand{\thefootnote}{\arabic{footnote}}

\section{Introduction}

Along with increasing concerns and stricter regulations~\cite{european_commission_regulation_2016} regarding data privacy, Federated Learning (FL)~\cite{mcmahan2017communication} has attracted widespread attention from both industry and academia.
As a paradigm for collaboratively training machine learning models without access to dispersed and private data, FL has been successfully used in many real-world applications including keyboard prediction~\cite{hard2018federated}, Internet of Things~\cite{khan2021federated}, and medical image analysis~\cite{li2020multi}. However, with the popularity of FL, we also face severe security threats from potential adversaries in real-world scenarios. Focusing on breaching the availability of FL models, poisoning attacks~\cite{baruch2019little,tolpegin2020data, shejwalkar2021manipulating} could completely corrupt the FL models. Focusing on breaching the confidentiality of FL models, inference attacks~\cite{melis2019exploiting, nasr2019comprehensive} and reconstruction attacks~\cite{zhu2019deep} could significantly increase the privacy risk of FL models. To eliminate the above threats, many defense methods~\cite{blanchard2017machine,shejwalkar2022back,huang2021evaluating} have been proposed and achieved outstanding defense performance against these attacks. 

Meanwhile, due to the distributed nature of FL system, the backdoor attack is emerging and becoming one of the most serious security threats to FL system~\cite{bagdasaryan2020backdoor, sun2019can,wang2020attack, goldblum2022dataset,kairouz2021advances}.
Backdoor attacks aim only to mislead backdoored models to exhibit abnormal behavior on samples with backdoor triggers. Similar to poisoning attacks, the adversary manipulates some training samples, such as adding a small patch in samples or adding extra specific samples, to insert a backdoor trigger into the model. Compared with poisoning attacks that aim to corrupt FL models' prediction performance or make FL training diverge~\cite{fang2020local,blanchard2017machine,baruch2019little,tolpegin2020data}, backdoor attacks could only be activated by data with triggers so that they are more stealthy and much harder to be detected and defended against. How to effectively defend against backdoor attacks to FL methods is still an open problem~\cite{bagdasaryan2020backdoor,sun2019can,wang2020attack,goldblum2022dataset}. Existing defense methods that perform well against poisoning attacks like Krum~\cite{blanchard2017machine}, norm clipping~\cite{shejwalkar2022back}, and adding noise~\cite{Du2020Robust,wang2020attack} do not achieve expected results against backdoor attacks~\cite{wang2020attack} (also seeing results in Section~\ref{sec:overall_evaluation}). Backdoor attacks in FL framework could be separated into two categories: white-box and black-box attacks~\cite{goldblum2022dataset,lyu2020privacy}. Compared with white-box attacks which allow adversarial clients to control the local training process, adversarial clients are only allowed to manipulate local training datasets under the black-box setting. Therefore, black-box attacks with few knowledge requirements are more practical for real-world scenarios~\cite{shejwalkar2022back} and also lead to non-negligible risks to FL systems~\cite{wang2020attack,shejwalkar2022back} (seeing results in Section~\ref{sec:overall_evaluation}).

A recent trend in FL deployment is to utilize personalized FL (pFL) methods~\cite{tan2022towards,kairouz2021advances,chen2022pfl,chen2023pFedGate} to mitigate the data heterogeneity issue across clients, a universal characteristic inherent in all real-world datasets. Being different from general FL methods only having a global model, pFL methods allow each client to have their own local personalized models, by finetuning~\cite{liang2020think,zhang2022fine}, interpolating~\cite{li2021ditto,fallah2020personalized}, or partially sharing the global model~\cite{li2021fedbn,collins2021exploiting}. This brings better adaptability on local private datasets and helps pFL methods surpass general FL methods by a large margin on prediction accuracy under practical Non-IID scenarios~\cite{chen2022pfl,wu2022motley}. 

The unique aspect of pFL, where clients have their own personalized local models, is a major contributor to its success. This distinction prompts us to consider whether the difference between clients' local models could prevent the injection and spread of backdoor features. Specifically, in this study, we aim to determine whether pFL methods can also provide robustness against difficulty-eradicated backdoor attacks. We ask the below questions:

\textbf{\textit{Do pFL methods bring robustness again backdoor attacks? And if the answer is yes, what leads to robustness, and how to utilize it?}}

To answer the above questions, we study the robustness of mainstreamed pFL methods against backdoor attacks in this paper. Our work focuses on black-box backdoor attacks as they efficiently exploit security risks in more practical settings. We test $4$ widely used backdoor attacks against $6$ pFL methods on benchmark datasets FEMNIST and CIFAR-10, a total of 600 experiments. 

For the first question, we find that various pFL methods show different robustness against backdoor attacks. \textbf{Some pFL methods (partial model-sharing) could achieve outstanding defense performance}, especially FedRep~\cite{collins2021exploiting} could reduce backdoor attack success rate below $10\%$. Even compared with some widely used defense methods, they are able to achieve similar and better robustness while still keeping excellent prediction accuracy. 

Through a series of experiments to dig out the source of robustness from pFL methods, we find that \textbf{the degree of personalization is a key factor to robustness benefits}. We observe that pFL methods with full model-sharing do not improve robustness against backdoor attacks. In contrast, pFL methods with partial model-sharing, FedBN~\cite{li2021fedbn}, and FedRep~\cite{collins2021exploiting}, significantly improve defense performance. These observations suggest a strong positive correlation between robustness against backdoor attacks and the larger personalization degree of pFL methods. Then, to further analyze the reasons behind different robustness from pFL methods, we conduct ablation studies on pFL methods. For full model-sharing methods, we find that if training of local models is more dependent on the global model, they are more vulnerable to backdoor attacks. For partial model-sharing methods like FedBN or FedRep which allow each client to own their locally preserved BN layers or linear classifiers, we find that they successfully block the propagation of backdoor features between personalized local models.

Finally, inspired by our findings, \textbf{we further propose a defense method, \textit{Simple-Tuning}}, that allows training FL models while efficiently reducing vulnerability to backdoor attacks. Simple-Tuning first reinitializes the linear classifier of trained models of each client and then trains it on local training datasets. As only tuning the linear classifier, Simple-Tuning is easier to be combined with existing FL methods with significantly reduced computation costs. We test defense performance of Simple-Tuning, and the results show that it significantly boosts robustness against backdoor attacks.

\paragraph{\textbf{Contributions.}} We summarize our contributions as follows:

\noindent\textbf{(1)} We present the first detailed and thorough study that tests the robustness of 6 popular pFL methods against $4$ widely used backdoor attacks. We find that some pFL methods (partial model-sharing) could achieve outstanding defense performance against backdoor attacks.

\noindent\textbf{(2)} We find that the degree of model-sharing is a key factor to robustness benefits from pFL methods. pFL methods with full model-sharing do not show robustness against backdoor attacks. In contrast, pFL methods with partial model-sharing, FedBN~\cite{li2021fedbn}, and FedRep~\cite{collins2021exploiting}, significantly improve defense performance against backdoor attacks.

\noindent\textbf{(3)} We also compare pFL methods with some widely used defense methods in FL against backdoor attacks. pFL methods are able to achieve similar and even better robustness while still keeping excellent prediction accuracy.

\noindent\textbf{(4)} We conduct ablation studies on pFL methods to further analyze the reasons behind various robustness performances of pFL methods.

\noindent\textbf{(5)} Based on our findings, we propose a lightweight and plug-and-play defense method, Simple-Tuning. The experiments show its effectiveness in boosting FL models' robustness against backdoor attacks. 

\section{Background}
\label{sec:backgrounds}
\subsection{Personalized Federated Learning}
\label{sec:pfl_methods}
Before conducting our studies, we first introduce Federated Learning (FL) and personalized Federated Learning (pFL). Under FL scenario, each client $i\in \mathcal{C}$ has his own private dataset $\mathcal{S}_{i}$ drawn from its own local data distribution $\mathcal{D}_{i}$ over $\mathcal{X}\times \mathcal{Y}$. The goal of FL is to train a single (global) model collaboratively without sharing clients' local data. The objective of FL is 
\begin{equation}
\label{eq:fl_formulation}
\begin{aligned}
\min _{\bm{\theta}_g} & \  \mathbb{E}_{(\bm{x}, \bm{y}) \sim \mathcal{D}_i,i \in \mathcal{C}} \left[ \mathcal{L}(f(\bm{\theta}_g; \bm{x}), \bm{y})\right]
\end{aligned}
\end{equation}
where $\bm{\theta}_{g} $ is the global model shared with all clients. $\mathcal{L}$ is the loss function, such as cross-entropy of the classification task. FedAvg~\cite{mcmahan2017communication} is the most widely used method to solve Eq.\ref{eq:fl_formulation}. For each round of FL training, several clients are selected by the server to participate in the training. The server first broadcasts $\bm{\theta}_{g}$ to selected clients. All clients then locally train $f(\bm{\theta}_{g}; \cdot)$ on their private dataset by using SGD for several epochs and upload the update of $\bm{\theta}_{g}$ to the server. After receiving all clients' update, the server conducts the weighted average of updates by
\begin{flalign}
\label{eq:avg}
    \overline{\nabla \bm{\theta}_g} = \sum_{j=1}^{M} \frac{|\mathcal{S}_{j}|}{|\mathcal{S}|} \nabla \bm{\theta}_{g}^{j}, \ \ |\mathcal{S}| = \sum_{j=1}^{M} |\mathcal{S}_{j}|.
\end{flalign}
Here, $|\mathcal{S}_{j}|$ is the size of dataset owned by the selected client $j$. The server applies the $\overline{\nabla \bm{\theta}_{g}}$ into $\bm{\theta}_{g}$ for the next round of FL training, and the process repeats until convergence. 

Being different from FL, pFL methods not only tend to learn a global model but also aim to learn a local personalized model for each client. The local personalized model is better adaptive to each client's local dataset and helps pFL methods surpass general FL methods by a large margin on prediction accuracy under practical Non-IID scenarios~\cite{tan2022towards,hanzely2021personalized,pillutla22a,collins2021exploiting,liang2020think,smith2017federated,zhang2022fine}. Here, \textbf{we use below formulations to illustrate various pFL methods based on different personalization degrees}. 

\paragraph{\textbf{(i)} Full model-sharing~\cite{li2021ditto,liang2020think,t2020personalized,marfoq2021federated,fallah2020personalized}:}

\begin{equation}
\label{eq:full-model-sharing}
\begin{aligned}
\min _{\bm{\theta}_{g}, \{\bm{\theta}_i\}_{i\in C}} & \mathbb{E}_{(\bm{x}, \bm{y}) \sim \mathcal{D}_i,i \in \mathcal{C}} \left[ \mathcal{L}(f(\bm{\theta_i}; \bm{x}), \bm{y}) \right] + \lambda \mathcal{H} (\bm{\theta}_{g}, \bm{\theta}_{i}),\\ 
\end{aligned}
\end{equation}
where $\bm{\theta}_{i}$ represents local personalized model owned by each client and $\bm{\theta}_{g}$ means the global reference model shared among all clients. We can observe that the whole global model $\bm{\theta}_{g}$ is shared across all clients. 
The new added $\mathcal{H}$ is the regularizer of similarity between $\bm{\theta}_{g}$ and $\bm{\theta}_{i}$ and $\lambda$ is the coefficient for each client. In each round, selected clients receive $\bm{\theta}_{g}$ and use it as the reference to training $\bm{\theta}_{i}$. $\bm{\theta}_{g}$ is also updated on local datasets and then uploaded to the server to participate in the aggregation process. Therefore, full model-sharing also indicates knowledge from every client's local dataset could be transferred to local models of other clients by sharing $\bm{\theta}_{g}$.

Here, we introduce several pFL methods with full model-sharing used in our experiments. \textbf{(1)} \textit{Ditto}~\cite{li2021ditto}: each client has two models, global model $\bm{\theta}_{g}$ and local model $\bm{\theta}_{i}$. During local training, after receiving $\bm{\theta}_g$, clients first train $f(\bm{\theta}_g;\cdot)$ on their local datasets to update $\bm{\theta}_g$. Then they train local model $\bm{\theta}_i$ using $\bm{\theta}_{g}$ as the reference with a added regularizer $\lambda \|\bm{\theta}_i-\bm{\theta}_{g}\|_2$. Here, $\lambda$ controls the consistence between $\bm{\theta}_{g}$ and $\bm{\theta}_{i}$, the personalization degree of each client. \textbf{(2)} \textit{pFedMe}~\cite{t2020personalized}: By first setting $\bm{\theta}_{g}$ as the reference, clients utilize the Moreau envelop similar to \textit{Ditto} regularizer to update $\bm{\theta}_{i}$. Then they update $\bm{\theta}_{g}$ by using aggregation of the updates of $\bm{\theta}_{i}$ at the end of each local epoch. \textbf{(3)} \textit{FedEM}~\cite{marfoq2021federated}: this method assumes the data distribution of each client is the mixture of unknown underlying data distributions with different coefficients. Therefore, they propose to learn a mixture of multiple base models ($a_i^1\bm{\theta}^{1}+\dots+a_i^k\bm{\theta}^{k}, ~ i \in C$) for each client with different mixture coefficients. All clients share these based models $\bm{\theta}^{1}, \dots, \bm{\theta}^{k}$ but they have their own personalized mixture coefficients, $a_i^1, \dots, a_i^k$. \textbf{(4)} \textit{Fine-tuning}: It is a widely used method to obtain a personalized model for each client and sometimes shows a better performance than other pFL methods~\cite{chen2022pfl,matsuda2022empirical}, in which each client obtains $\bm{\theta}_{i}$ by Fine-tuning $\bm{\theta}_{g}$ on their local datasets after FL training finishes.  

\paragraph{\textbf{(ii)} Partial model-sharing~\cite{pillutla22a,collins2021exploiting,li2021fedbn,arivazhagan2019federated}:}
\begin{equation}
\label{eq:partial-model-sharing}
\min_{\bm{\theta}^{s},\left\{\bm{\theta}_i^p\right\}_{i\in C}} \mathbb{E}_{(\bm{x}, \bm{y}) \sim \mathcal{D}_i,i \in \mathcal{C}} \left[ \mathcal{L}(f(\bm{\theta}^s, \bm{\theta}_i^p; \bm{x}), \bm{y}) \right],
\end{equation}
where the local personalized model $\bm{\theta}_{i}$ is further separated into two group parameters, $\bm{\theta}^{s}$ and $\bm{\theta}^p_{i}$. There is no more global model and only partial parameters $\bm{\theta}^{s}$ of $\bm{\theta}_{i}$ are shared across clients. After finishing local training, global shared parameters $\bm{\theta}^{s}$ are uploaded to the server for aggregation. Locally parameters $\bm{\theta}^p_i$ are never shared with other clients. Partial model-sharing indicates each client could only utilize knowledge of partial model parameters trained on the other clients' local datasets so that each client could gain larger personalization degrees. For example, 
\textbf{(5)} \textit{FedBN}~\cite{li2021fedbn} decouples the whole model $\bm{\theta}_{i}$ into locally preserved BN layers $\bm{\theta}^p_{i}$ and shared parameters $\bm{\theta}^s$. After the local training process finishes, clients only upload $\bm{\theta}^s$ to participate in aggregation. \textbf{(6)} \textit{FedRep}~\cite{collins2021exploiting,pillutla22a} decouples models into global feature extractor ($\bm{\theta}^{s}$) and local linear classifier ($\bm{\theta}_{i}^{p}$). In each iteration of local training, clients first freeze the feature extractor and update  $\bm{\theta}_{i}^p$ on their private datasets. And then, they freeze the local linear classifier and update global feature extractor $\bm{\theta}^{s}$. Only the global feature extractor is uploaded to the server and participates in aggregation.

\subsection{Backdoor Attacks in FL}
\label{sec:background_backdoor_attacks}

\label{sec:backdoor_attacks}
\begin{figure}
    \centering
    \includegraphics[width=0.45\textwidth]{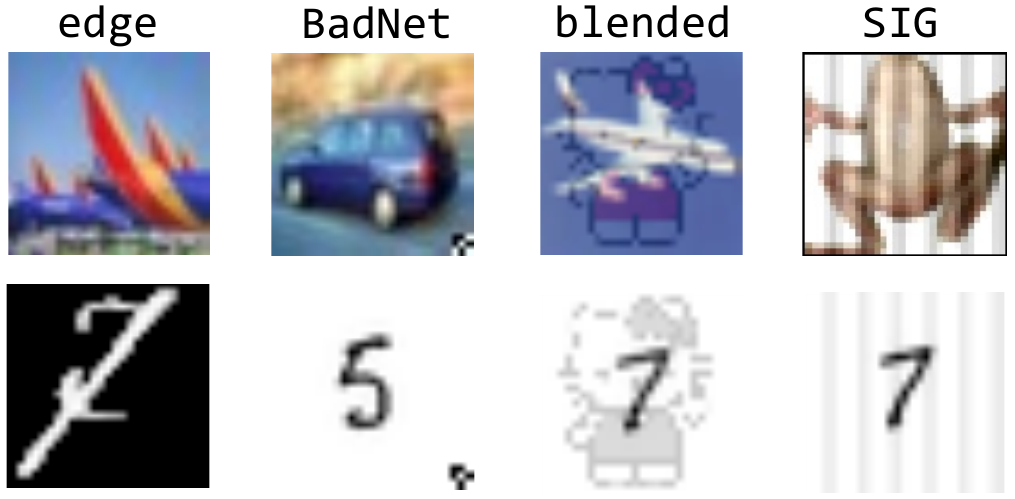}
    \caption{We show used backdoor triggers: edge-case, BadNet, Blended, SIG triggers. The first and second row are visualizations on CIFAR-10 and FEMNIST, respectively.}
    \label{fig:backdoor_images}
\end{figure}

Backdoor attacks aim to mislead the backdoored model to exhibit abnormal behavior on samples stamped with the backdoor trigger but behave normally on all benign samples. As shown in Figure~\ref{fig:backdoor_images}, the backdoor attacker inserts a backdoor trigger into the model by manipulating some training data samples, like adding a small patch in clean images or just choosing special samples as triggers. Backdoored samples' labels would also be changed to attacker designated target label $\bm{y}_t$. In the training phase, the model $f(\bm{\theta}'; \cdot)$ is trained on clean data samples $(\bm{x}, \bm{y})$ together with backdoored samples $(\bm{x}', \bm{y}_t)$. During the inference phase, the presence of a backdoor trigger in input samples will mislead backdoored model $f(\bm{\theta}'; \cdot)$ to predict the target label $\bm{y}_t$ while keeping the same performance on samples without the trigger.  Therefore, compared with poisoning attacks which corrupt FL models' prediction performance or make FL training diverge~\cite{fang2020local,blanchard2017machine,baruch2019little,tolpegin2020data}, backdoor attacks are more stealthy and hard to detect since the backdoored model behaves normally in the absence of triggers~\cite{goldblum2022dataset}. 

Backdoor Attacks in FL could be separated into 2 categories: \textbf{(1)} white-box setting where the adversarial client can control the whole local training process and local updates to the server; \textbf{(2)} black-box setting where the adversarial client is only allowed to manipulate his own local dataset. In the black-box setting, the adversary does not require computing resources and any access to the internal configurations (e.g., FL binaries, memory) of compromised devices. Therefore, backdoor attacks in black-box setting is much more practical in real-world scenarios~\cite{shejwalkar2022back,lyu2020privacy,kairouz2021advances} and also more widely used in previous works of backdoor learning~\cite{wang2020attack,shejwalkar2022back,wu2022backdoorbench}. 

Bagdasaryan et al~\cite{bagdasaryan2020backdoor} propose the ﬁrst backdoor attack in FL. One or multiple adversarial clients use several special training samples as semantic triggers which are inconsistent with other images in the same class. The authors also assume that attackers always conduct attacks near the convergent phase of FL training.
Wang et al~\cite{wang2020attack} propose edge-case backdoor triggers by choosing samples from other data distributions as backdoor triggers. They think these samples are located in the tail of the original data distribution. Therefore, the model updates from those samples are unlikely to conflict with model updates from other benign clients. Their experiments also show that edge-case triggers are more effective and persistent than semantic triggers. Besides, the authors also use a more practical attack setting, where the attacker could only periodically participates in several FL rounds, for example, in every fixed $Q$ round. 

In DBA attack~\cite{Xie2020DBA}, the authors propose to use the BadNet~\cite{gu2019badnets} trigger which is the most widely used attack in centralized training. Besides, DBA attack also assumes the attacker can control multiple adversarial clients. Each adversary injects a different local trigger into his own dataset, trains the model on this backdoored dataset, and uploads its model updates to the server. During the inference process, the attacker uses the combination of these local triggers as a new trigger to conduct the attack.

Besides, other works also propose white-box backdoor attacks~\cite{bagdasaryan2020backdoor,bhagoji2019analyzing,zhang2022neurotoxin,Xie2020DBA}, which allow attackers to control the local training process, such as model replacement attack to magnify poisoning update~\cite{bagdasaryan2020backdoor}. We leave evaluations about them in future work.   

\section{Threat Model}
\label{sec:attack_setting}
In this work, we focus on \textbf{black-box backdoor attacks} that do not require control and knowledge about training procedures like model architecture, parameters, or training methods. They also do not require any computing resources which makes them essential threat in real-world scenarios. For simplicity, we consider the task of image classification that has been used in most research on backdoor attacks.

In our studies, we choose the stricter assumption that there is only one adversarial client. Following~\cite{wang2020attack,sun2019can}, we adopt the fix-frequency attack setting in which the adversarial client is selected to join in the training process for every fixed $Q$ round, where we set $Q=10$ through all experiments. 

We adopt commonly used \textbf{edge-case} and \textbf{BadNet} triggers in our evaluations and also explore various trigger types, including \textbf{Blended}~\cite{chen2017targeted} and \textbf{SIG}~\cite{barni2019new} triggers, which are widely used in centralized training but not yet studied in FL. Blended trigger extends BadNet by encouraging the invisibility of triggers through blending clean images and triggers. We choose the hello-kitty (hk) pattern as our trigger since it achieves the best attack performance in related backdoor defense works~\cite{huang2022backdoor,wu2021adversarial,wu2022backdoorbench,zheng2022data}. Without changing image pixels, SIG adopts a sinusoidal signal as the trigger to perturb clean images. Examples of backdoor triggers  are listed in Figure~\ref{fig:backdoor_images}.

\section{Robustness of pFL methods against Backdoor Attacks}

In this work, we want to learn that apart from better prediction accuracy, do pFL methods bring robustness again backdoor attacks? If the answer is true, what factors of pFL methods lead to robustness? And based on our studies, could we exploit this robustness benefit to help improve the robustness of FL models against backdoor attacks? We first try to answer the first question by conducting the evaluations of backdoor attacks on pFL methods mentioned before. We outline our experimental setup in Section \ref{sec:setting} and show overall experimental results in Section \ref{sec:overall_evaluation} followed by our interesting findings. We also compare some widely used defense methods in FL in Section \ref{sec:overall_evaluation}. 

To further analyze the reasons behind various robustness
from pFL methods, we conduct ablation studies on pFL methods with partial model-sharing in Section \ref{sec:partial_analysis} and methods with full model-sharing in Section \ref{sec:full_analysis}, respectively. In Section \ref{sec:new_defense}, based on previous findings, we propose a simple defense method, \textit{Simple-Tuning} and further verify its effectiveness of defending against backdoor attacks by taking experiments on FL methods.

\subsection{Experimental Settings}
\label{sec:setting}

\subsubsection{Datasets and Models} 
We conduct evaluations on two widely used image classification datasets, \textbf{FEMNIST}~\cite{caldas2018leaf} and \textbf{CIFAR-10}~\cite{krizhevsky2009learning} in FL literature. 
FEMNIST is a handwritten character recognition dataset containing 62 classes and each client corresponds to a character writer from EMNIST\cite{cohen2017emnist}. 
Following pFL-bench\cite{chen2022pfl}, we adopt the sub-sampled version of FEMNIST in our experiment, and it contains 200 clients. And the dataset is randomly split into train/valid/test sets with a ratio 3:1:1. For CIFAR-10, following previous works~\cite{chen2022pfl, caldas2018leaf, wang2020attack, zawad2021curse}, we use Dirichlet allocation to split 
it into 100 clients with Dirichlet factor $\alpha=0.5$.
These two datasets represent two different settings in Non-IID setting: feature-skew and label skew~\cite{chen2022pfl,lyu2020privacy,zawad2021curse}. 
Besides, for CIFAR-10, we also consider the IID setting in which we uniformly and randomly sample images from each class for each client. The splitting details about datasets are shown in Section \ref{sec:data-model} in \textit{Appendix}.

Following pFL works~\cite{chen2022pfl,liang2020think,t2020personalized}, we use a simple ConvNet model on these two datasets. The details are shown in Section \ref{sec:data-model} in \textit{Appendix}. Besides, to align with previous backdoor attacks works~\cite{wang2020attack,zhang2022neurotoxin}, we also use a larger network, ResNet-18~\cite{he2016deep}, for CIFAR-10 dataset. 

\subsubsection{Personalized FL Methods.}
We first consider the baseline FL method, \textit{FedAvg}~\cite{mcmahan2017communication}. We adopt mainstreamed pFL methods mentioned in Section \ref{sec:pfl_methods}, \textit{Fine-tuning}~\cite{liang2020think}, \textit{FedBN}~\cite{li2021fedbn}, \textit{FedRep}~\cite{collins2021exploiting,pillutla22a}, \textit{Ditto}~\cite{li2021ditto}, \textit{pFedMe}~\cite{t2020personalized}, \textit{FedEM}~\cite{marfoq2021federated}. They are also used in recent benchmark~\cite{chen2022pfl,matsuda2022empirical}. We implement these methods following pFL-bench \footnote{\url{https://github.com/alibaba/FederatedScope/tree/master/benchmark/pFL-Bench}}~\cite{chen2022pfl} and defer the choice of hyperparameters in Section \ref{sec:implementation_details} in \textit{Appendix}.

\subsubsection{Training Details.} The total training round $T$ is set to be $1,000$. For each training round, the server randomly samples $10\%$ clients from all clients to participate in the training process. That is to say, in each round, $20$ and $10$ clients are randomly selected for FEMNIST and CIFAR-10 dataset.

\subsubsection{Attack Settings.} 
Through our evaluations, for simplicity, we choose the client-$1$ and client-$15$ as the adversarial client on FEMNIST and CIFAR-10, respectively. We set the target label $\bm{y_t} =1$ on FEMNIST and $\bm{y_t} = 9$ on CIFAR-10 following~\cite{wang2020attack}. For FEMNIST dataset, we use 100 images of “7”s from Ardis~\cite{kusetogullari2020ardis} as edge-case triggers. For CIFAR-10 dataset, we use 500 images of Southwest Airline’s planes provided in~\cite{wang2020attack} as edge-case triggers. For each of BadNet, Blended, and SIG, we put it on $50\%$ of training samples of the adversarial client. Following attack hyparameters provided in~\cite{wu2022backdoorbench}, for BadNet attack, we choose the $3\times3$ grid pattern and put it at the downright corner of poisoned samples. For Blended attack, we choose the hello-kitty pattern and set the blending ratio $\alpha$ as $0.2$. For SIG attack, we set the amplitude of sinusoidal signal as 20 ($[0,255]$). We also show these used triggers in Figure \ref{fig:backdoor_images}.

\subsubsection{Baseline Defense Methods.} We adopt four baseline defense methods in FL: norm-clipping~\cite{wang2020attack,shejwalkar2022back,sun2019can}, adding noise~\cite{wang2020attack,sun2019can,Du2020Robust}, Krum~\cite{blanchard2017machine}, and Multi-Krum~\cite{blanchard2017machine} which are widely used in previous works~\cite{bagdasaryan2020backdoor,sun2019can,wang2020attack,zhang2022neurotoxin,shejwalkar2022back,Du2020Robust,bagdasaryan2020backdoor,Xie2020DBA}. \textbf{(1)} Krum only selects one uploaded model update which is similar to other updates by computing parameter similarity ($\ell_{2}$ norm) between them as the aggregation result. Rather than only selecting one update, Multi-Krum selects top-k updates based on ranking of computed similarity and then takes an average as the aggregation result. \textbf{(2)} Norm-clipping: Norm-clipping clips the model update so as not to exceed the given threshold $c$ before aggregation. We set $c=1$ and $c=0.5$ following~\cite{zhang2022neurotoxin}, where a smaller $c$ means the smaller contribution to model aggregation from local model update. \textbf{(3)} Adding noise method adds Gaussian noises $\lambda*\bm{v}$ to model updates from clients. $\bm{v}$ is sampled from normal distribution. We try a series of noise scales $\sigma =10^{-3}, 5*10^{-4}$, and $10^{-4}$. We implement them based on source code \footnote{\url{https://github.com/ksreenivasan/OOD_Federated_Learning}} of~\cite{wang2020attack}.

\subsubsection{Evaluation Details.} 
We take two evaluation metrics, including \textit{Clean Accuracy (C-Acc)} (i.e., the prediction accuracy of clean samples) and \textit{Attack Success Rate (ASR)} (i.e., the prediction accuracy of poisoned samples to the target class). We test C-Accs on all clients' testing dataset using their own pFL models. For edge-case attack, we follow~\cite{wang2020attack} to test ASR on $100$ edge-case images on FEMNIST and $196$ edge-case images on CIFAR-10, respectively. For BadNet, Blended, and SIG attacks, we test ASR on testing sets of all clients except the selected adversarial client. All experiments are conducted with $3$ times over different random seeds, and we report the average through experiments.

\subsection{The Overall Robustness Evaluation }
\label{sec:overall_evaluation}

\begin{figure*}
    \centering
     \begin{subfigure}[b]{\textwidth}
         \centering
         \includegraphics[width=\textwidth]{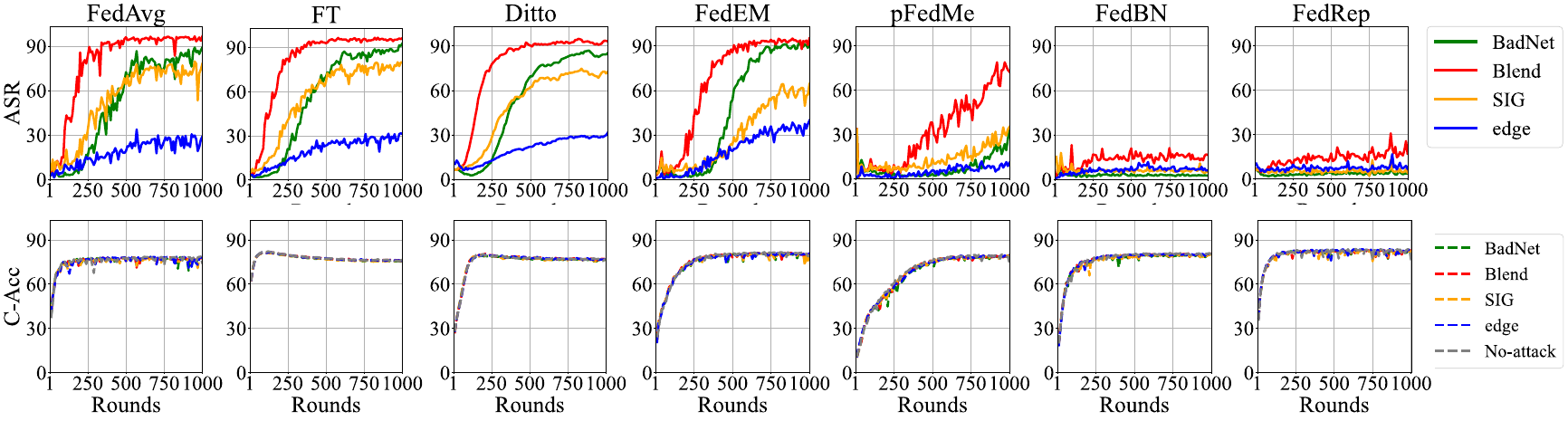}
         \caption{ResNet-18@CIFAR-10}
         \label{fig:resnet18_cifar10}
     \end{subfigure}
     \vfill
     \begin{subfigure}[b]{\textwidth}
         \centering
         \includegraphics[width=\textwidth]{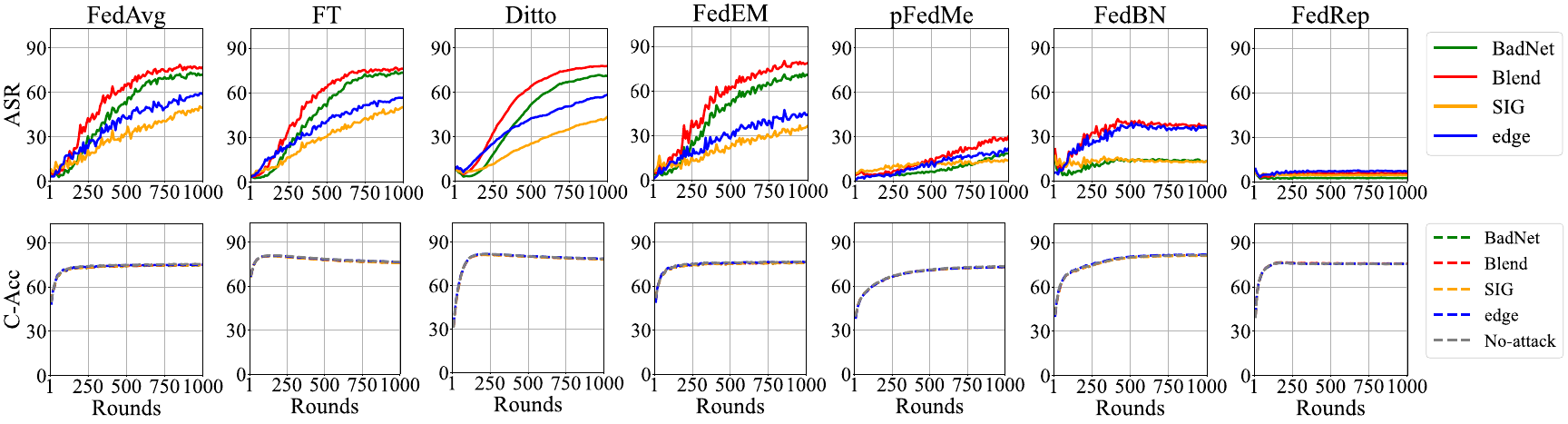}
         \caption{ConvNet@CIFAR-10}
         \label{fig:convnet2_cifar10}
     \end{subfigure}
     \vfill
     \begin{subfigure}[b]{\textwidth}
         \centering
         \includegraphics[width=\textwidth]{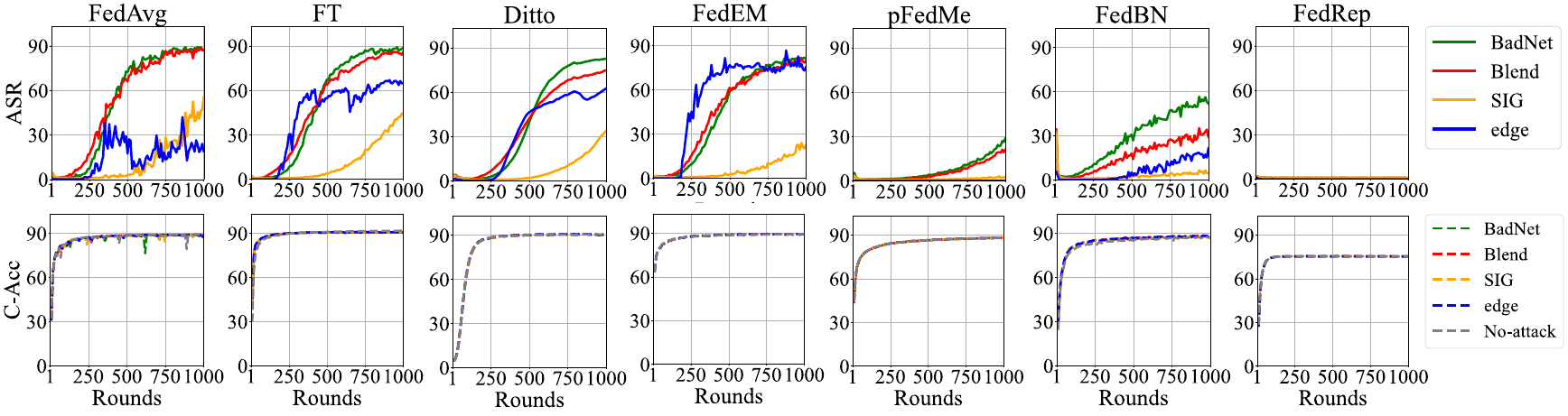}
         \caption{ConvNet@FEMNIST}
         \label{fig:convnet2_femnist}
     \end{subfigure}
     \caption{Comparison of backdoor attacks on different pFL methods. (a): Results on ResNet-18 and CIFAR-10; (b) Results on ConvNet and CIFAR-10; (c) Results on ConvNet and FEMNIST. In each figure, the value of horizontal or vertical axis is the number of training rounds or attack success rate, respectively. The solid line means ASR, and the dashed line means the C-Acc.}  \label{fig:overall_evaluation}
\end{figure*}

In Figure~\ref{fig:overall_evaluation}, we show performance curves of backdoor attacks on pFL methods under Non-IID data distribution. Due to the space limitation, the results of IID setting are deferred in Section \ref{sec:missed_results} of $\textit{Appendix}$. There are three subfigures: results on ResNet-18 and CIFAR-10, results on ConvNet and CIFAR-10, and results on ConvNet and FEMNIST. In each subfigure, we first demonstrate the backdoor attack success rate (ASR) and then the original task's clean accuracy (C-Acc). Each column represents different FL and pFL methods. The solid line represents ASR, and the dashed line represents the C-Acc. 

\paragraph{\textbf{Black-box backdoor attacks achieve good attack performance.}} We first look at how black-box backdoor attacks perform on pFL methods. From Figure~\ref{fig:overall_evaluation}, we first observe that even without requiring access to the FL training process, simple black-box backdoor attacks especially BadNet and Blended attacks achieve outstanding performance. This shows that practical black-box backdoor attacks are non-negligible threats. And, for all pFL methods, backdoor attacks do not affect C-Acc. As shown in each subfigure, different color dash lines almost overlap with the grey dash line (clean accuracy without attacks). It also reflects the stealthiness of black-box backdoor attacks that behave normally on samples without triggers. Then, among these four attack methods, Blended attack (the solid red line) achieves the best attack performance on CIFAR-10 dataset with over $90\%$ ASR on two models. 
On FEMNIST, BadNet attack achieves the best attack performance, followed by Blended attack. 

\paragraph{\textbf{pFL methods with partial model-sharing effectively alleviate backdoor attacks}} Next, we can clearly observe that pFL methods with partial model-sharing (last two columns), FedRep and FedBN, show outstanding robustness against backdoor attacks. FedRep shows the best defense performance against attacks and can limit the success rate of Blended attack and other attacks below $20\%$ and $10\%$. FedBN achieves the second-best defense performance on CIFAR-10 dataset and reduces ASR of four attacks below $20\%$ on the larger model, ResNet-18. These positive results demonstrate that \textbf{apart from improving prediction accuracy, some pFL methods (partial model-sharing) also bring better robustness against backdoor attacks.} 
 
\paragraph{\textbf{The degree of personalization is a key factor to robustness
benefits.}} Across columns, we observe that pFL methods with partial model-sharing significantly improve defense performance, but all pFL methods with full model-sharing (from the 2nd to 5th column) do not show improvement in the robustness against backdoor attacks except for pFedMe method. These observations suggest a strong positive correlation between robustness against backdoor attacks and the larger personalization degree of pFL methods. In Section~\ref{sec:analysis_model_sharing}, we will deeply investigate why different model-sharing degrees lead to various defensive effects and show that pfedMe is also vulnerable to backdoor attacks like other full model-sharing methods.  

Besides, comparing results of two different models on CIFAR-10 dataset, we could observe that backdoor attacks achieve better attack performance on high-capacity model (ResNet-18) than on the low-capacity model (ConvNet), which indicates that larger models may be more vulnerable to be injected with backdoor triggers in FL. The similar trend is also observed in~\cite{wang2020attack,wu2022backdoorbench}.

\subsection{The Robustness Evaluation of Baseline Defense Methods}

\begin{figure}
    \centering
    \includegraphics[width=0.47\textwidth]{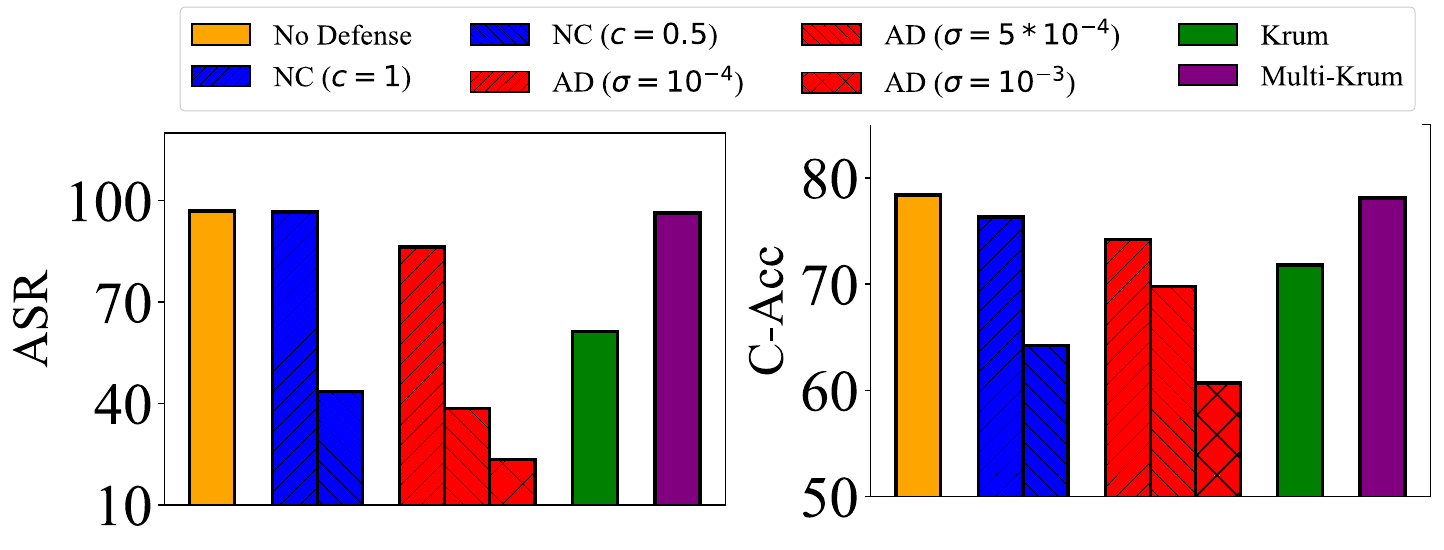}
    \vspace{-0.2cm}
    \caption{The evaluation of FedAvg with baseline defense methods against Blended attack on ResNet-18. Left: The ASR of Blended attack; Right: The C-Acc of FedAvg with defense strategies.}
    \label{fig:defense}
\end{figure}

To further compare the robustness brought by pFL methods, we also conduct experiments to evaluate defense performance of some widely used defense strategies in FL. We adopt four defense methods: norm-clipping, adding noise, Krum, and Multi-Krum. The hyperparameters have been outlined in Section \ref{sec:setting}. In~\cite{wang2020attack}, these four defense methods, especially Krum and Multi-Krum, show robustness against edge-case attack. Here, we implement these defense methods following~\cite{wang2020attack} and test them on the Blended attack which behaves better performance in the above studies. We still adopt the Non-IID setting and choose the FedAvg training method. The final epoch's ASR and C-Acc are shown in Figure~\ref{fig:defense}.

We first observe the defense performance of the simple norm clipping (NC) method. NC with $c=1$ used in~\cite{wang2020attack} couldn't bring any robustness improvement against the Blended attack. We further reduce the norm constrain of NC on local update to $0.5$. While reducing ASR of Blended attack, it also causes a significant drop on C-Acc. Like norm clipping, with increased strength of added noise $\sigma$, AD significantly reduces the ASR of Blended attack. However, it also leads to a significant drop of C-Acc. Similar to performance on the edge-case attack, Krum also improves the robustness against Blended attack. However, since only one model would be selected in every aggregation, the C-Acc also experiences a significant drop. Rather than only selecting one model update as the result of aggregation, Multi-Krum selects top-$k$ ($k=7$) model updates and then averages updates in the aggregation. Although Multi-Krum achieves better C-Acc, it fails to defend against Blended attack. Although NC, AD, and, Krum show different robustness improvements against Blended attack, \textbf{they all face a serious trade-off between robustness and clean accuracy}. 

\section{Analysis of Various Robustness from pFL Against Backdoor Attacks}
\label{sec:analysis_model_sharing}

In Section~\ref{sec:overall_evaluation}, we observe that pFL methods with partial model-sharing, FedRep, and FedBN, gain much better robustness against backdoor attacks than full model-sharing methods. In following sections, we take detailed studies of pFL methods to analyze the reasons behind various robustness from them. We first look at FedBN and FedRep in Section \ref{sec:partial_analysis} and then analyze full model-sharing methods in Section 
\ref{sec:full_analysis}.

\begin{figure}[t!]
    \centering
    \includegraphics[width=0.35\textwidth]{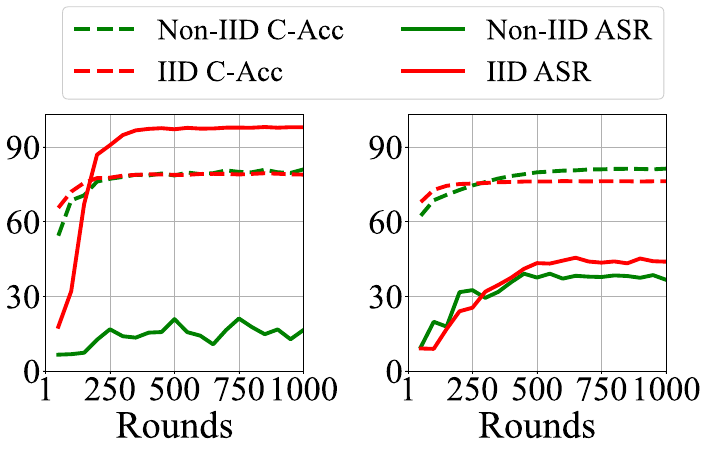}
    \vspace{-0.2cm}
    \caption{The results of Blended attack on FedBN on CIFAR-10 dataset under Non-IID and IID setting. Left: On ResNet-18; Right: On ConvNet.}
    \label{fig:fedbn_iid}
\end{figure}

\subsection{pFL Methods with Partial model-sharing}
\label{sec:partial_analysis}

\paragraph{FedBN} 
In FedBN, as every client shares all parameters except local BN layers, data heterogeneity among clients leads to differences in local BN layers' parameters. We think differences in local BN layers across clients block backdoor feature propagation in local models.
For simplicity, each local model could be denoted as a neural network $F_{i}^{(l)}$ with $l$ layers, 
$F_i^{(l)}=f^{(l)} \circ \phi_i \circ f^{(l-1)} \circ \phi_i \circ \cdots \circ f^{(1)}$,
where $i$ is the client index and $l$ is total number of layers. $\phi_i$ are local BN layers and the remaining $f$ are shared parameters $\bm{\theta}^s$. Even if backdoor features have been learned by $\bm{\theta}^{s}$, due to different $\phi_i$ from FedBN, neurons corresponding to backdoor features after the BN layers of other local models won't be activated by triggers. Therefore, FedBN could perform better robustness against backdoor attacks. To verify this, we conduct evaluations of Blended attack on FedBN under Non-IID and IID settings on CIFIAR-10. 
According to our previous analysis, under IID setting, without data heterogeneity, local BN layers of clients become more consistent with each other, which should lead to better attack results under this setting. As shown in Figure \ref{fig:fedbn_iid}, ASR of IID is higher than that of Non-IID. Especially on ResNet-18, FedBN does not show a defensive effect, which is consistent with our previous analysis.

\begin{figure}
    \centering
    \includegraphics[width=0.45\textwidth]{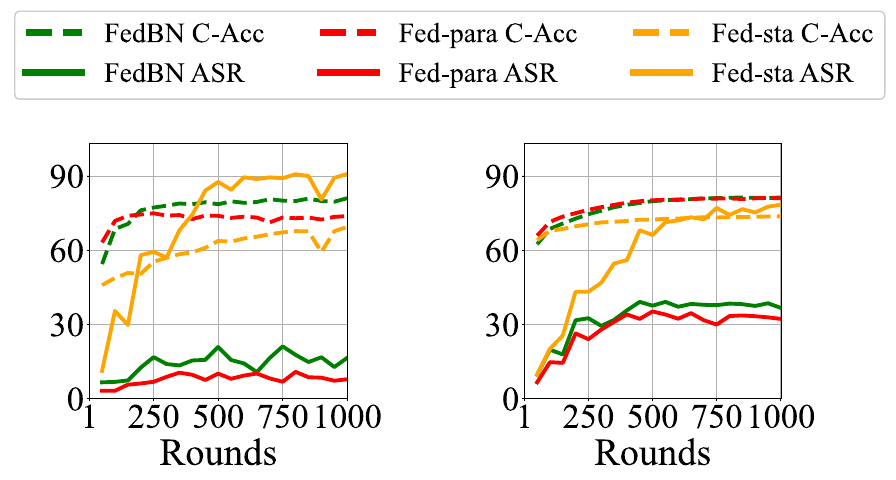}
    \vspace{-0.3cm}
    \caption{The evaluation of each part of BN layers. Left: On ResNet-18; Right: On ConvNet.}
    \label{fig:fedbn_aba}
\end{figure}

To further analyze defense performance of local BN layers, we take attack evaluations on each part of BN layers, i.e., running statistics $\mu$, $\sigma^2$ and learnable parameters $\gamma$, $\beta$. 
Unlike FedBN, we only choose not to share running statistics or learnable parameters. We denote them as \textit{Fed-sta} or \textit{Fed-para}. The comparison between Fed-sta, Fed-para, and the original FedBN on the Blended attack and CIFAR-10 dataset is shown in~Fig \ref{fig:fedbn_aba}. We can observe that in terms of C-Acc, Fed-sta and Fed-para experience a drop on ResNet-18 but the latter drops less. In ConvNet, Fed-para keeps the same C-Acc as FedBN. Surprisingly, we find that Fed-para achieves even better robustness than FedBN. In contrast, Fed-sta performs the worst defensive effect against Blended attack, and ASR even increases to $90\%$ on ResNet-18. Those results demonstrate the differences in learnable parameters may be more important in preventing the propagation of backdoor features than the differences in running statistics of BN layers. 

\begin{figure}
    \centering
    \includegraphics[width=0.44\textwidth]{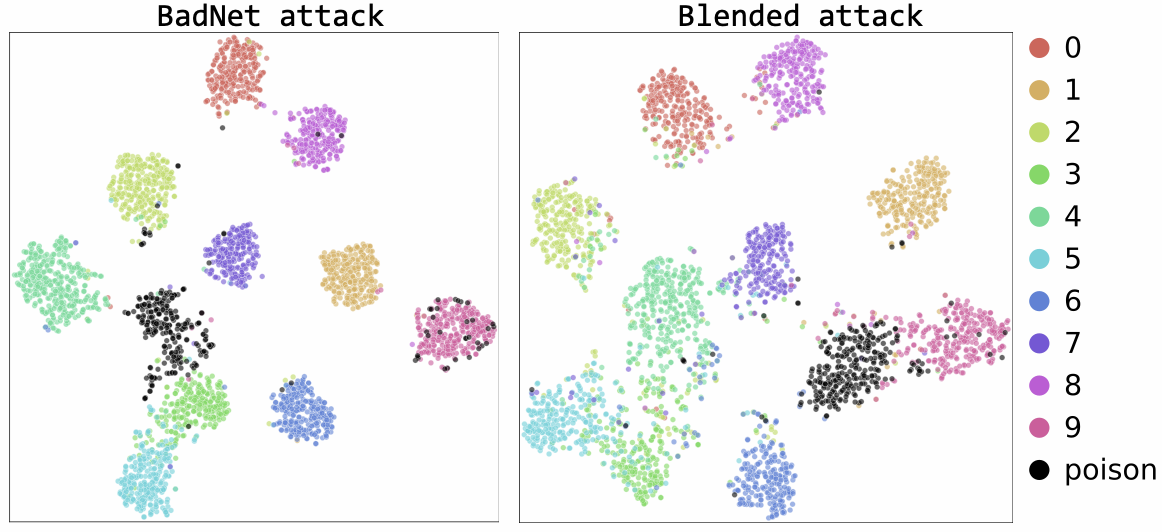}
    \vspace{-0.2cm}
    \caption{The T-SNE visualization on feature space of global feature extractor of FedRep. Each color denotes each class, and the black points represent backdoored samples.}
    \label{fig:feature-visualization}
\end{figure}

\paragraph{FedRep} 
FedRep decouples the local model into global feature extractor $\bm{\theta}^{s}$ and local linear classifier $\bm{\theta}^{p}_{i}$ and alternatively trains them on local datasets. And only feature extractor $\bm{\theta}^{s}$ is updated and shared across clients. We would like to know which part of FedRep leads to or contributes most to improving robustness against backdoor attacks. We first study the global feature extractor $\bm{\theta}^{s}$ to figure out whether it has learned backdoor triggers even under an alternative training procedure. We conduct feature visualization on feature space of $\bm{\theta}^{s}$ by using T-SNE~\cite{van2008visualizing}. Specifically, we randomly choose $4$ benign clients with the adversarial client and visualize their training samples in feature space for both BadNet and Blended attacks. We take experiments on ResNet-18 and CIFAR-10. The results are shown in Figure \ref{fig:feature-visualization}, where black points are backdoored samples. We could observe that most backdoored samples gather together in a single cluster, which indicates that the global feature extractor has already learned backdoor features. 

Therefore, FedRep's outstanding robustness against backdoor attacks comes from the local linear classifier $\bm{\theta}^{p}_{i}$ of each client. $\bm{\theta}^{p}_{i}$ of benign clients are updated only based on their own clean dataset and never shared with other clients. Therefore, benign client's linear classifiers $\bm{\theta}^{p}_{i}$  do not learn the mapping from backdoor features to target label $\bm{y}_{t}$. Inspired by this exciting finding, we further propose a simple defense method and will discuss it in Section \ref{sec:new_defense}.  

\subsection{pFL Methods with Full model-sharing}
\label{sec:full_analysis}

\paragraph{Ditto.} 

\begin{figure}
    \centering
\includegraphics[width=0.325\textwidth]{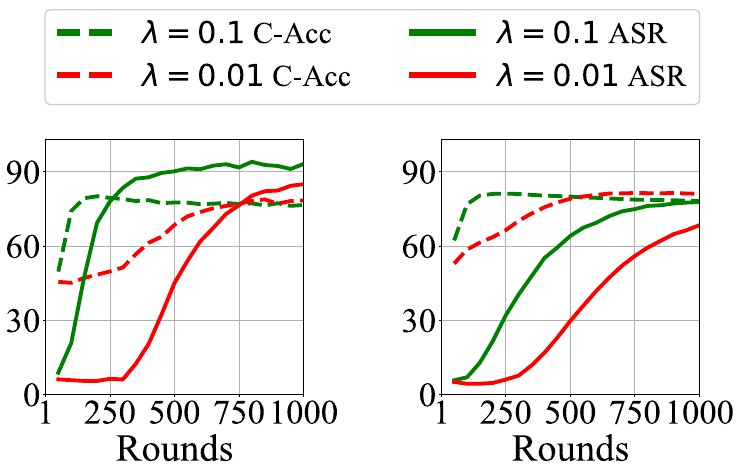}
    \vspace{-0.3cm}
    \caption{The results of Blended attack on Ditto with different $\lambda$ on CIFAR-10 dataset. Left: ResNet-18; Right: ConvNet.}
    \label{fig:ditto}
\end{figure}

Backdoor features could be inserted into $\bm{\theta}_{g}$ during local training on the adversary client and then propagated to other clients by full model-sharing. In Ditto~\cite{li2021ditto}, clients use the global model $\bm{\theta}_{g}$ as the reference to guide training of local model $\bm{\theta}_{i}$ based on $\lambda \mathcal{H} (\bm{\theta}_{g}, \bm{\theta}_{i})$. $\lambda$ controls the similarity with $\bm{\theta}_{g}$ and also the personalization degree of $\bm{\theta}_{i}$. The smaller $\lambda$ means the $\bm{\theta}_{i}$'s update is less dependent on $\bm{\theta}_{g}$. Since previous results indicate a strong positive correlation
between robustness against backdoor attacks and the larger personalization degree, we investigate if the smaller $\lambda$ (larger personalization degree) can block the transfer of backdoor features from $\bm{\theta}_{g}$ to $\bm{\theta}_{i}$. We conduct experiments of Blended attack on CIFAR-10 dataset with $\lambda = 0.1$ and $\lambda=0.01$ in Figure~\ref{fig:ditto}. We could observe that with smaller $\lambda$, ASR drops slightly on ResNet-18 and ConvNet. This verifies our thought. However, in terms of clean accuracy, the convergence of Ditto also significantly slows down. Therefore, Ditto also faces a trade-off between backdoor robustness and clean accuracy of the main task like mentioned defense methods.

\begin{figure}
    \centering
\includegraphics[width=0.32\textwidth]{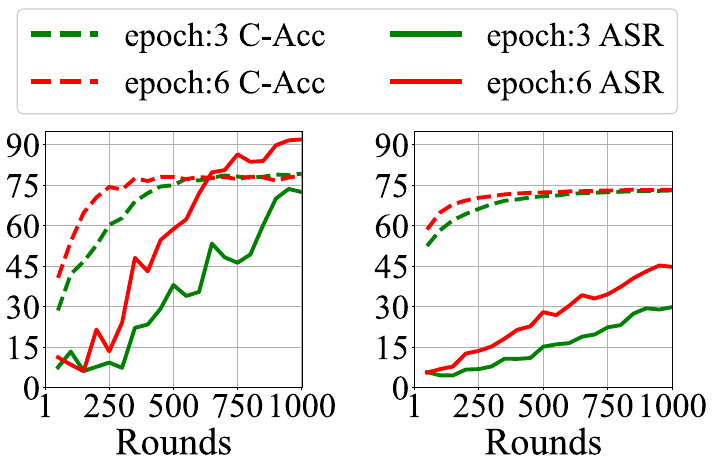}
    \vspace{-0.3cm}
    \caption{The results of Blended attack on pFedMe with the different number of local training epoches on CIFAR-10. Left: on ResNet-18; Right: on ConvNet.}
    \label{fig:pfedme_aba}
\end{figure}

\paragraph{pFedMe} Similar to Ditto, pFedMe~\cite{t2020personalized} also trains local personalized model of each client with a penalty term $\lambda \mathcal{H} (\bm{\theta}_{g}, \bm{\theta}_{i})$ but performs a better defense performance against backdoor attacks. The reason is that pFedMe only updates the global model at the end of each epoch of local training based on local model updates, which leads to fewer chances for backdoor triggers to insert into the global model by using pFedMe method. Compared with Ditto which updates $\bm{\theta}_{g}$ at each batch in local training, few updates of $\bm{\theta}_{g}$ on the adversarial client hinder the injection of backdoor signatures. Therefore, we test Blended attack performance on ResNet-18 and ConvNet on CIFAR-10 while increasing the number of local epochs from the original $3$ to $6$. The results are shown in Figure~\ref{fig:pfedme_aba}. We can observe that ASR increases by $15\%$ along with increasing local epoch number. This demonstrates that pFedMe is also vulnerable to backdoor attacks like Ditto.

\begin{figure}
    \centering
\includegraphics[width=0.455\textwidth]{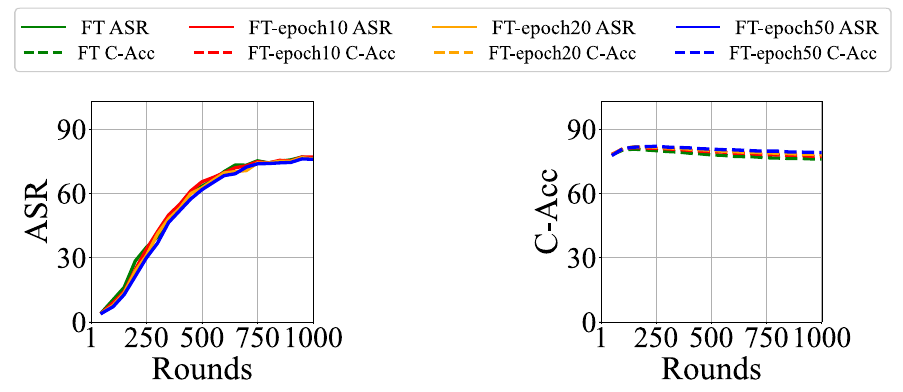}
    \vspace{-0.3cm}
    \caption{The results of Blended attack for FT with various epoch numbers on ConvNet and CIFAR-10.}
    \label{fig:ft_abl}
\end{figure}

\paragraph{FedEM} In FedEM~\cite{marfoq2021federated}, each client trains multiple base models on local datasets, and all base models are later uploaded to aggregate in the server. Therefore, backdoor features could be learned by the base models from adversarial clients and are later propagated to other clients by full model-sharing.

\paragraph{Fine-tuning.} Fine-tuning (FT) is also a widely used backdoor defense method in centralized setting~\cite{liu2018fine,wu2022backdoorbench,huang2022backdoor,li2021neural} which further updates trained model to forget backdoor triggers. However, we find it couldn't improve the robustness under FL setting. Specifically, we set the epoch number of FT as $5$, $10$, $20$, and $50$ to observe if the robustness will increase with the increase of FT epochs. The results on ConvNet and CIFAR-10 are shown in Figure \ref{fig:ft_abl}. Even if we use 50 epochs to conduct local Fine-tuning,  ASR remains unchanged. The first potential reason is that there are no sufficient clean training examples for Fine-tuning to purify the backdoored global model. Training samples owned by each client are much less than training samples under centralized setting, which restricts the effect of Fine-tuning. Besides, it is also observed that FT couldn't effectively reduce ASR of backdoor attacks with low poisoning rate~\cite{wu2022backdoorbench}. Since there is only one adversarial client and $50\%$ of training samples of the adversarial client are backdoor samples in our evaluations, the proportion of poisoned samples in all training samples is still very low with only $0.55\%$ (275/50000).

\begin{table}[t]
\caption{The results on ResNet-18 and ConvNet models on CIFAR-10 dataset. We adopt trained models from FedAvg and Ditto methods. A lower ASR means better robustness. A higher ACC means better C-Acc. The best results are highlighted in \textbf{bold}.}
\vspace{-0.3cm}
\scalebox{0.76}{
\begin{tabular}{c|c|cc|cc|cc}
\hline
\multirow{2}{*}{Models}       & Methods & \multicolumn{2}{c|}{original}     & \multicolumn{2}{c|}{+ FT-linear}  & \multicolumn{2}{c}{+ ST}        
\\ \cline{2-8} 
& Attacks & \multicolumn{1}{c}{ASR($\downarrow$)}  & Acc($\uparrow$) & \multicolumn{1}{c}{ASR($\downarrow$)}  & Acc($\uparrow$)& \multicolumn{1}{c}{ASR($\downarrow$)}  & Acc($\uparrow$)\\ \hline
\multirow{2}{*}{Res18@FedAvg} & BadNet  & \multicolumn{1}{c}{89.2} & 78.1  & \multicolumn{1}{c}{90.1} & 76.8  & \multicolumn{1}{c}{\textbf{39.3}} & \textbf{80.9}  \\ 
& Blended & \multicolumn{1}{c}{97.5} & 78.1  & \multicolumn{1}{c}{97.4} & 76.7  & \multicolumn{1}{c}{\textbf{47.2}} & \textbf{81.1}  \\ \hline

\multirow{2}{*}{Conv@FedAvg}  & BadNet  & \multicolumn{1}{c}{76.5} & 74.4  & \multicolumn{1}{c}{71.0} & 74.8  & \multicolumn{1}{c}{\textbf{6.4}} & \textbf{76.0}  \\ 
& Blended & \multicolumn{1}{c}{76.7} & 74.5  & \multicolumn{1}{c}{77.1} & 74.7  & \multicolumn{1}{c}{\textbf{13.8}} & \textbf{76.1}  \\ \hline

\multirow{2}{*}{Res18@Ditto} & BadNet  & \multicolumn{1}{c}{85.1} & 76.5  & \multicolumn{1}{c}{86.2} & 74.5  & \multicolumn{1}{c}{\textbf{38.9}} & \textbf{80.8}  \\ 
& Blended & \multicolumn{1}{c}{95.6} & 76.6  & \multicolumn{1}{c}{95.7} & 74.4  & \multicolumn{1}{c}{\textbf{47.1}} & \textbf{81.2}  \\ \hline

\multirow{2}{*}{Conv@Ditto}  & BadNet  & \multicolumn{1}{c}{71.2} & \textbf{77.9}  & \multicolumn{1}{c}{72.6} & 76.5  & \multicolumn{1}{c}{\textbf{5.6}} & 76.7 \\ 
& Blended & \multicolumn{1}{c}{76.8} & \textbf{77.8}  & \multicolumn{1}{c}{77.1} & 76.6  & \multicolumn{1}{c}{\textbf{17.4}} & 76.8  \\ \hline
\end{tabular}}
\vspace{-0.2cm}
\label{tabel_tuning_exp}
\end{table}

\section{Improving Backdoor Robustness with Simple-Tuning}
\label{sec:new_defense}

In previous sections, we observed that pFL methods with partial model-sharing could achieve better robustness against backdoor attacks. By conducting detailed studies on them in Section~\ref{sec:partial_analysis}, we found that locally trained BN layers in FedBN or locally trained linear classifiers in FedRep plays the key roles in improving robustness. Therefore, it is natural to consider, whether can we provide some insight for designing better robust FL methods to further boost robustness based on our findings and analyses? In this section, we take an initial step by providing a simple but effective defense method.  

 Based on the idea of locally training a linear classifier of FedRep, we propose \textit{Simple-Tuning} which only tunes the linear classifier of FL models of each client after the training process. Specifically, we first \textbf{reinitialize} the linear classifier and \textbf{retrain} it with the local training dataset of each client, while freezing the remaining parameters of each model. Compared with vanilla Fine-tuning, Simple-Tuning (ST) has made two improvements: \textbf{(1)} Compared with tuning the whole model in FT, we only tune the linear classifier after finishing the training process, which could efficiently reduce computation costs; \textbf{(2)} Rather than inheriting parameters of original models like FT, we choose to reinitialize and retrain local linear classifier on local datasets.

We test our method on trained models from FedAvg and Ditto and show defense performances against BadNet and Blended attacks. For Simple-Tuning, we adopt default Kaiming Uniform normalization~\cite{he2015delving} and use the constant learning rate as $0.005$. We only tune the linear classifier for $10$ epochs. The results containing ASR and C-Acc on ResNet-18 and ConvNet models are shown in Table~\ref{tabel_tuning_exp}. We compare the results of our method with the results of original FedAvg and Ditto models and models from only fine-tuning local linear classifiers without reinitialization. We denote the latter as FT-linear. As shown in Table~\ref{tabel_tuning_exp}, our method, Simple-Tuning, significantly improves robustness against backdoor attacks. Compared with original models, it efficiently reduces the ASR of two backdoor attacks by $56.6\%$ on average. It achieves better robustness on smaller model ConvNet, reducing ASR below $20\%$. Surprisingly, Simple-Tuning even improves the C-Acc of FL methods except for Ditto on ConvNet. These results demonstrate the great potential of Simple-Tuning. It is worth noting that FT-linear does not show any robustness like vanilla FT. It also verifies the importance of reinitialization in our method to backdoor robustness. This suggests that directly fine-tuning FL backdoor models may not efficiently remove backdoor triggers. It also contrasts with findings of \cite{adi2018turning} which utilizes FT methods to purify watermark, specific backdoor trigger, in DNN models. However, they demonstrated that all three methods - vanilla FT, FT-linear, and FT-linear with randomized linear classifier - can effectively purify watermark triggers. We believe this is mainly because triggers of the watermark are set differently from the general backdoor triggers. In \cite{adi2018turning}, the authors set $\bm{y}_t$ of each watermarked sample to a randomly chosen label, rather than a fixed target label as in backdoor attacks.
 
By only tuning the linear classifier, Simple-Tuning is easier to be combined with existing FL methods. It is also more convenient to deploy in real-world scenarios with significantly reduced computation costs. However, Simple-Tuning still couldn't completely eliminate security risks from backdoor attacks. We hope it could inspire future work to propose better defense mechanisms, such as using more advanced optimizers and designing better FT methods.

\section{Conclusion}

This paper studies the robustness of popular personalized FL methods against backdoor attacks in FL. We conduct the first study of backdoor attacks in the pFL framework, testing 4 widely used backdoor attacks against 6 pFL methods on benchmark datasets FEMNIST and CIFAR-10, a total of 600 experiments. We show that pFL methods with partial model-sharing achieve outstanding robustness against backdoor attacks. Based on our in-depth ablation studies on various pFL methods, we find there is a strong positive correlation between robustness and a larger personalization degree. Inspired by our findings, we further propose a simple defense method that could effectively alleviate backdoor attacks with much fewer computational costs. We make two
categories of conclusions:

\paragraph{Next steps for researchers} We have shown that FedRep and our Simple-Tuning achieve good defense performance. It is an interesting question if there is a new way to design stronger backdoor attacks that reduce the effectiveness of FedRep and bypass our Simple-Tuning. It could help us better foresee potential security risks. 
Another direction is to design a better defense method suitable for FL settings. We should consider both defense performance and requirements of real-world scenarios like limited computational resources. 

\paragraph{Next steps for practitioners} The security threat, black-box backdoor attacks, we expose is immediately practical and lead to non-negligible risks. Therefore, this vulnerability will need to be considered during the deployment of FL models. Both improving prediction accuracy under Non-IID setting and boosting backdoor robustness, pFL methods with partial model-sharing may be better choices in real-world applications.

\section{Acknowledgements}

Zeyu Qin was supported by Alibaba Group through Alibaba Research Intern Program.

\bibliographystyle{ACM-Reference-Format}
\balance
\bibliography{sample-base}

\clearpage

\appendix

\renewcommand{\thepage}{A\arabic{page}}  
\renewcommand{\thesection}{A\arabic{section}}   
\renewcommand{\thetable}{A\arabic{table}}   
\renewcommand{\thefigure}{A\arabic{figure}}

\section{Datasets and Models} \label{sec:data-model}

\paragraph{Datasets.} We conduct experiments on two widely used dataset, FEMNIST and CIFAR-10 in FL literature~\cite{mcmahan2017communication, caldas2018leaf,chen2022pfl,matsuda2022empirical,wu2022motley,wang2020attack}. These two datasets represent two different settings of Non-IID respectively, feature-skew and label skew~\cite{chen2022pfl,lyu2020privacy,zawad2021curse}. 

\begin{figure}[]
    \centering
    \includegraphics[width=0.47\textwidth]{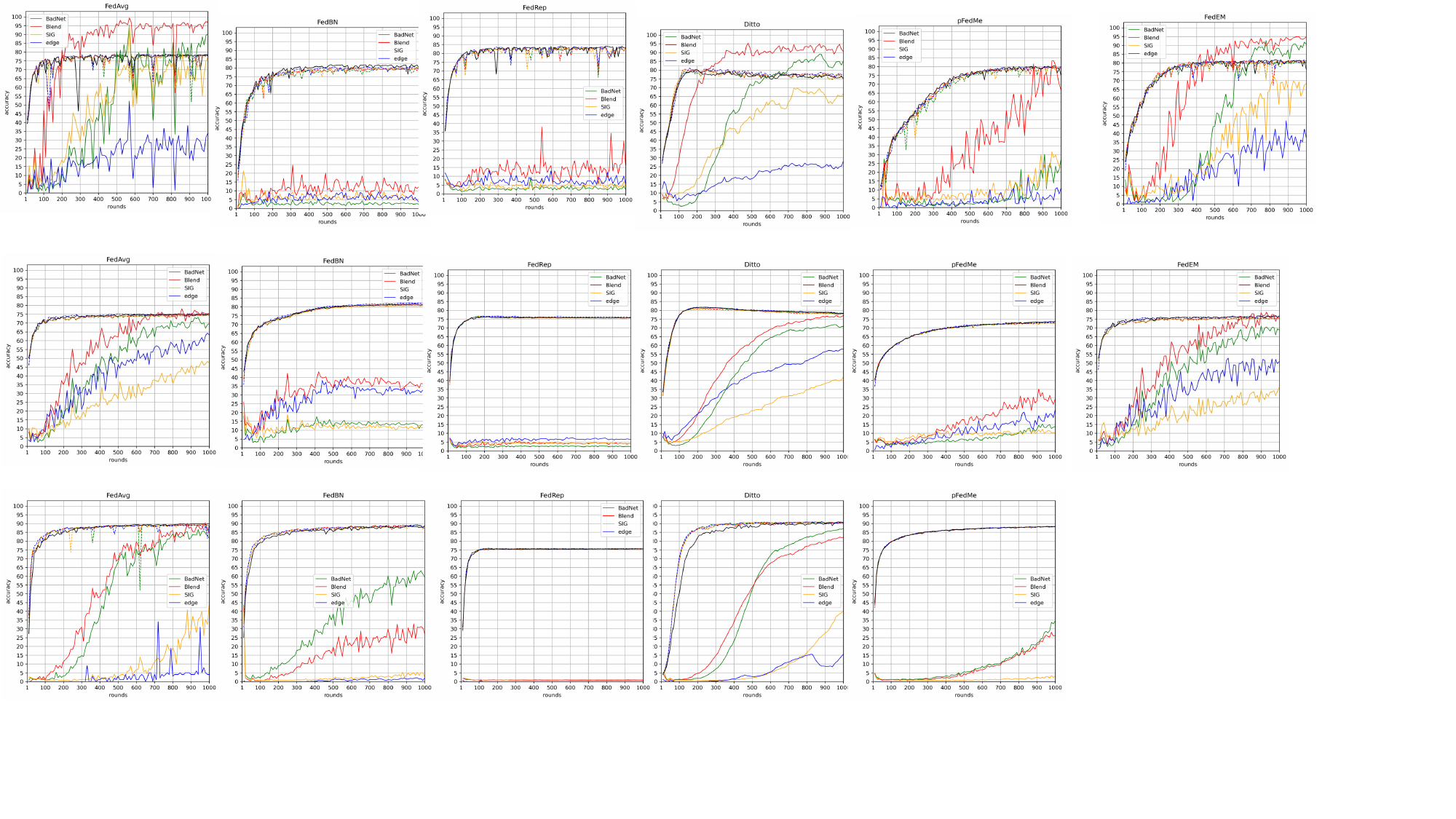}
    \caption{The violin plot of number of samples per client for CIFAR-10 and FEMNIST datasets.}
    \label{fig:vis_data_dis}
\end{figure}

\begin{itemize}
    \item FEMNIST: The Federated Extended MNIST (FEMNIST) is a widely used FL dataset for 62-class handwritten character recognition~\cite{caldas2018leaf}. The original FEMNIST dataset contains 3,550 clients and each client corresponds to a character writer from EMNIST\cite{cohen2017emnist}. We adopt the sub-sampled version in our evaluations, which contains 200 clients and totally $43,400$ images with resolution of $28\times28$ pixels.  The distribution of samples number per client is shown in Figure \ref{fig:vis_data_dis}. 
    \item CIFAR-10 is a 10-class image classiﬁcation dataset containing 60,000 colored images with resolution of 32x32 pixels. We use Dirichlet allocation to split this dataset into 100 clients with $\alpha = 0.5$. The distribution of samples number per client is shown in Figure \ref{fig:vis_data_dis}.  
\end{itemize}

\paragraph{Models.} Following the previous pFL works~\cite{t2020personalized,liang2020think,chen2022pfl,marfoq2021federated}, we utilize a simple ConvNet on FEMNIST and CIFAR10. This ConvNet model consists of two convolutional layers with the $5 \times 5$ kernel, max pooling, batch normalization, ReLU activation, and two dense linear layers. The hidden size of linear layer is $2,048$ and $512$ on FEMNIST and CIFAR10 respectively. To align with backdoor attacks works~\cite{wang2020attack,wu2022backdoorbench,sun2019can,zhang2022neurotoxin}, we also adopt the larger model, ResNet-18 on CIFAR-10 dataset.

\section{Implementation Details}
\label{sec:implementation_details}

\paragraph{Environment.} We implement our evaluations \footnote{\url{https://github.com/alibaba/FederatedScope/tree/backdoor-bench}} based on Pytorch~\cite{paszke2017automatic} and FederetadScope framework~\cite{federatedscope}. And all experiments are conducted on a cluster of 8 NVIDIA GeForce GTX 2080 Ti GPUs.

\subsection{More Details of Attacks Hyper-parameters}

Following attack hyparameters provided in~\cite{wu2022backdoorbench}, for BadNet attack, we choose the $3\times3$ grid pattern and put it at downright corner of poisoned samples. For Blended attack, we choose the hello-kitty pattern and set the blending ratio $\alpha$ as $0.2$. For SIG attack, we set the amplitude of sinusoidal signal as 20 ($[0,255]$).

\begin{figure*}[t]
    \centering
     \begin{subfigure}[b]{0.98\textwidth}
         \centering
         \includegraphics[width=\textwidth]{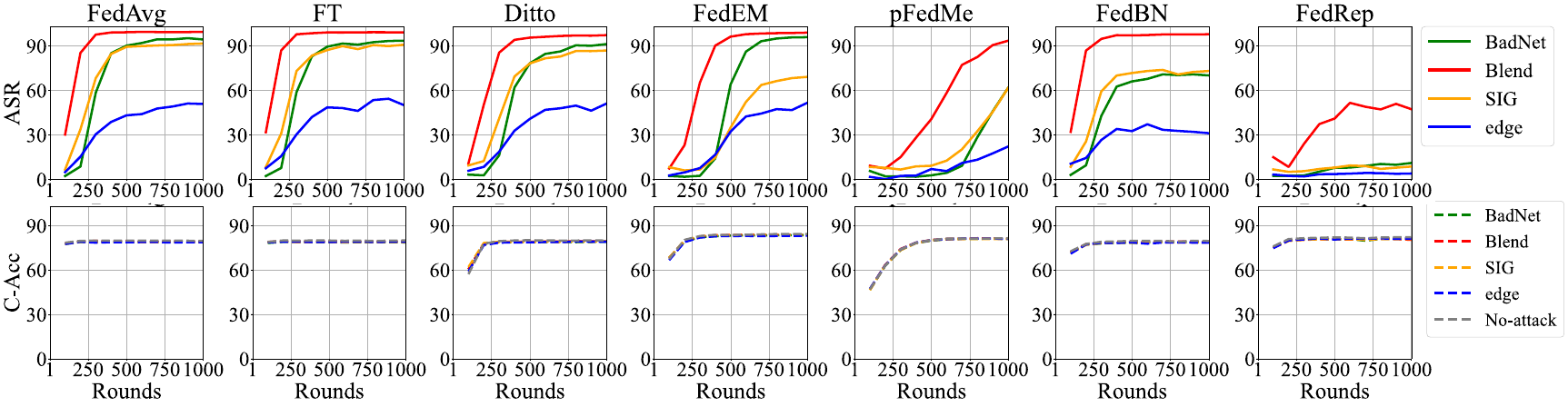}
         \caption{IID: ResNet-18@CIFAR-10}
         \label{fig:iid_resnet18_cifar10}
     \end{subfigure}
     \vfill
     \centering
     \begin{subfigure}[b]{0.98\textwidth}
         \centering
         \includegraphics[width=\textwidth]{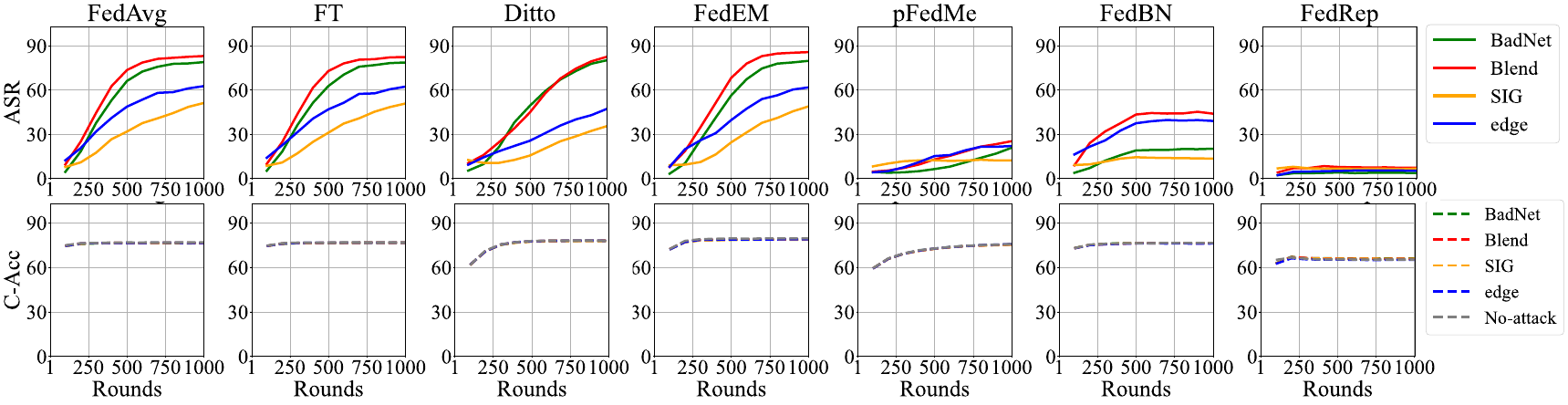}
         \caption{IID: Convnet2@CIFAR-10}
         \label{fig:iid_convnet2_cifar10}
     \end{subfigure}
     \caption{Comparison of backdoor attacks on different pFL methods. (a): Results on ResNet-18 and CIFAR-10; (b) Results on ConvNet and CIFAR-10. In the figures, the solid line means ASR, and the dashed line means the C-Acc.}
     \label{fig:iid_results}
\end{figure*}

\subsection{Implementation Details of pFL methods} 
Following pFL-bench \footnote{\url{https://github.com/alibaba/FederatedScope/tree/master/benchmark/pFL-Bench}}~\cite{chen2022pfl}, we also use hyper-parameter searching (HPO) algorithm, HyperBand~\cite{li2017hyperband}, to ﬁnd the best hyper-parameters for all the baselines on all datasets. We use the vanilla SGD optimizer and set the batch size of local datsets as $32$ for all methods. Here, we provide adopted hyper-parameters for all baseline methods:
\begin{itemize}
    \item FedAvg: we set learning rate as $0.1$ and epoch number as $2$ for each client on two datasets.
    \item Ditto: we set learning rate as $0.1$ and $\lambda$ as $0.1$ for each client. The epoch number is $2$ or $3$ on CIFAR-10 or FEMNIST. 
    \item FedEM:  ResNet-18 and CIFAR-10: we set learning rate as $0.5$ and epoch number as $1$. ConvNet and CIFAR-10 or FEMNIST: we set learning rate as $0.05$ and epoch number as $2$. The number of base model for two datasets is $3$.
    \item Fine-tuning: On CIFAR-10: we set learning rate of FT as $0.01$ and epoch number of FT as $2$. On FEMNIST: we set learning rate of FT as $0.01$ and epoch number of FT as $3$.
    \item pFedMe: On ResNet-18 and CIFAR-10: we set learning rate as $0.5$ and epoch number as $3$. We set local approximation steps $K$ as $1$, average moving parameter $\beta$ as $1$ and $\lambda$ as $0.5$;  On ConvNet and CIFAR-10: we set learning rate as $0.1$ and epoch number as $3$. We set local approximation steps $K$ as $2$, average moving parameter $\beta$ as $1$, and $\lambda$ as $0.5$; On ConvNet and FEMNIST: learning rate is $0.1$, epoch number is $2$, local approximation steps $K$  is $3$, average moving parameter $\beta$ is $1$, $\lambda$ is $0.8$.
    \item FedBN: On ResNet-18 and CIFAR-10: we set learning rate as $0.5$ and epoch number as $2$; On ConvNet and CIFAR-10: we learning rate as $0.1$ and epoch number as $2$; On ConvNet and FEMNIST: we set learning rate as $0.01$ and epoch number as $2$.
    
    \item FedRep: On CIFAR-10: we set learning rate of $\bm{\theta}^{s}$ as $0.1$, epoch number of $\bm{\theta}^{s}$ as $1$, learning rate of $\bm{\theta}^{p}$as $0.005$, and epoch number of $\bm{\theta}^{p}$ as $1$; On FENNIST: we set learning rate of $\bm{\theta}^{s}$ as $0.1$, epoch number of $\bm{\theta}^{s}$ as $2$, learning rate of $\bm{\theta}^{p}$ as $0.1$, and epoch number of $\bm{\theta}^{p}$ as $1$.
\end{itemize}

\section{Detailed Experimental Results}
\label{sec:missed_results}

\subsection{Experimental Results about IID Setting}

In this section, we demonstrate attack results on pFL methods under IID setting in which we uniformly and randomly sample images from each class of CIFAR-10 for each client. The results are shown in Figure~\ref{fig:iid_results}. We have similar observations as in the Non-IID scenario: \ding{182} Under IID setting, for all pFL methods, backdoor attacks do still not affect C-Acc. \ding{183} Under IID setting, Blended attack (the solid red line) also achieves the best ASR on all pFL methods and CIFAR-10 dataset. \ding{184} pFL methods with partial model-sharing also achieve better defense performance against backdoor attacks. Compared with attack results under Non-IID setting, attack results on FedBN under IID setting are higher. We have analysed reasons in Section~\ref{sec:partial_analysis}. \ding{185} Under IID scenario, we still observe that backdoor attacks achieves better attack performance on high-capacity model (ResNet-18) than on low-capacity model (ConvNet).

\end{document}